\documentclass[10pt,twocolumn,letterpaper]{article}

\usepackage{iccv}
\usepackage{times}
\usepackage{epsfig}
\usepackage{graphicx}
\usepackage{amsmath}
\usepackage{amssymb}
\usepackage{xcolor}
\usepackage[margin=0.1cm]{caption}

\usepackage[pagebackref=true,breaklinks=true,letterpaper=true,colorlinks,bookmarks=false]{hyperref}

\iccvfinalcopy 

\def\iccvPaperID{1479} 

\newcommand{\methodname}{{DeepProposal}}

\ificcvfinal\fi
\begin{document}

\title{\methodname: Hunting Objects by Cascading Deep Convolutional Layers}

\author{
Amir Ghodrati$^{1}$\thanks{A. Ghodrati and A.Diba contributed equally to this work},
Ali Diba$^{1}\footnotemark[1]$,
Marco Pedersoli$^{2}$\thanks{This work was carried out while he was in KU Leuven ESAT-PSI.}~\thanks{LEAR project, Inria Grenoble Rhˆone-Alpes, LJK, CNRS, Univ. Grenoble Alpes, France.},
Tinne Tuytelaars$^{1}$,
Luc Van Gool$^{1,3}$\\
$^{1}$KU Leuven, ESAT-PSI, iMinds \qquad  $^{2}$Inria  \qquad $^{3}$CVL, ETH Zurich \\
$^{1}${\tt\small firstname.lastname@esat.kuleuven.be} \quad $^{2}${\tt\small marco.pedersoli@inria.fr}\\
\and
}
\maketitle

\begin{abstract}
In this paper we evaluate the quality of the activation layers of a convolutional neural network (CNN) for the generation of object proposals. We generate hypotheses in a sliding-window fashion over different activation layers and
show that the final convolutional layers can find the object of interest with high recall but poor localization due to the coarseness of the feature maps. Instead, the first layers of the network can better localize the object of interest but with a reduced recall.
Based on this observation we design a method for proposing object locations that is based on CNN features and that combines the best of both worlds. We build an inverse cascade that, going from the final to the initial convolutional layers of the CNN, selects the most promising object locations and refines their boxes in a coarse-to-fine manner. The method is efficient, because i) it uses the same features extracted for detection, ii) it aggregates features using integral images, and iii) it avoids a dense evaluation of the proposals due to the inverse coarse-to-fine cascade. The method is also accurate; it outperforms most of the previously proposed object proposals approaches and when plugged into a CNN-based detector produces state-of-the-art detection performance.
\end{abstract}

\section{Introduction}
\label{sec:intro}

\begin{figure}
\begin{center}
{
\includegraphics[width=1\linewidth]{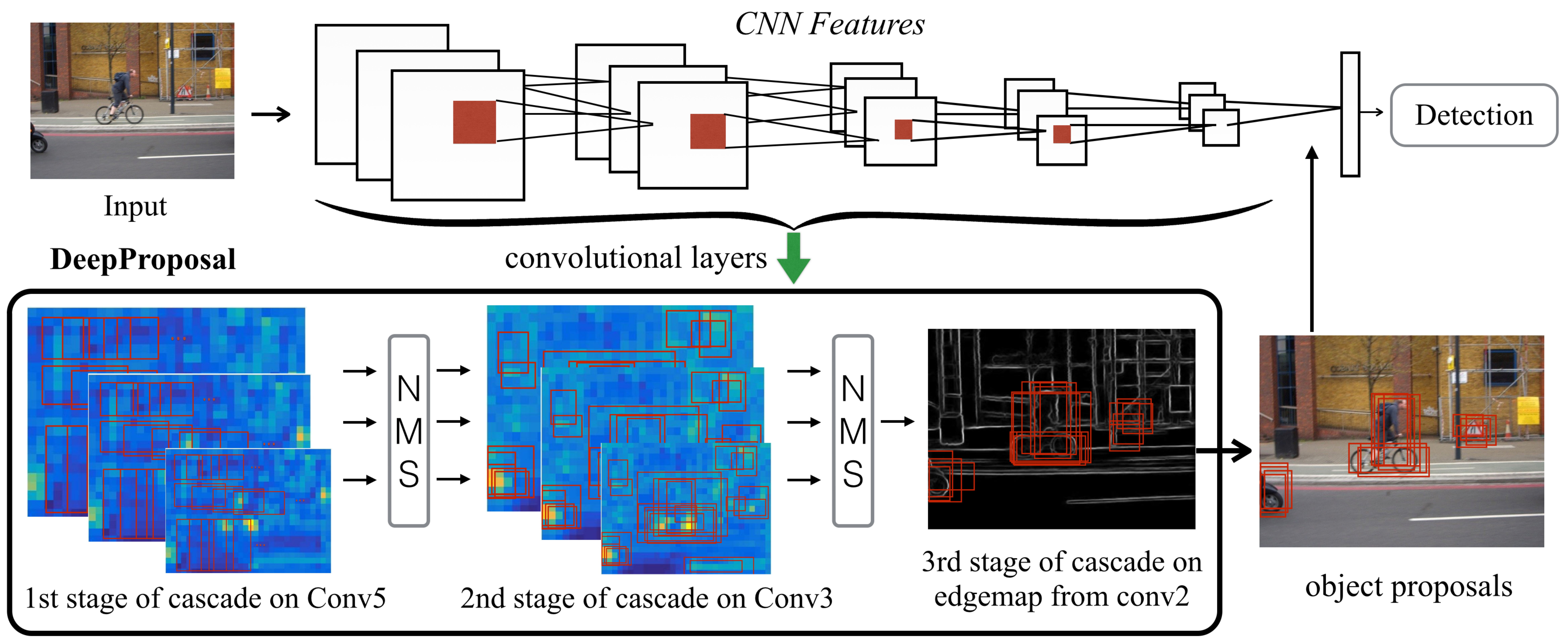}
}
\end{center}
\vspace{-0.4cm}
\caption{\methodname~object proposal framework. Our method uses deep convolutional layers features in a coarse-to-fine inverse cascading to obtain possible object proposals in an image. Starting from dense proposal sampling from the last convolutional layer (layer 5) we gradually filter irrelevant boxes until the initial layers of the net (layer 2). In the last stage we use contours extracted from layer 2, to refine the proposals. Finally the generated boxes can be used within an object detection pipeline.}
\label{fig:overview}
\end{figure}

In recent years, the paradigm of generating a reduced set of object location hypotheses (or window candidates) to be evaluated with a powerful classifier has become very popular in object detection. Most of the recent state-of-the-art detection methods \cite{segfisherDet,RCNN,sppnet,regionlets} are based on such proposals. Using limited number of these proposals also helps with weakly supervised learning, 

in particular learning to localize objects without any bounding box annotations~\cite{objlocalize,weaklydetection}.
This approach can be seen as a two-stage cascade: First, selection of a reduced set of promising and class-independent hypotheses and second, a class-specific classification of each hypothesis. This pipeline has the advantage that, similarly to sliding window, it casts the detection problem to a classification problem. However, in contrast to sliding window, more powerful and time consuming detectors can be employed as the number of candidate windows is reduced.

Methods for the generation of the window candidates are based on two very different approaches. The first approach uses bottom-up cues like image segmentation \cite{MCG,selectivesearch}, object edges and contours \cite{edgebox} for window generation. The second approach is based on top-down cues which learn to separate correct object hypotheses from other possible window locations \cite{objectness, BING}. So far, the latter strategy seems to have inferior performance. In this paper we show that, with the proper features, accurate and fast top-down window proposals can be generated.

We consider for this task the convolutional neural network (CNN) ``feature maps'' extracted from the intermediate convolutional layers of the Alexnet \cite{alexnet} trained on 1000 classes of ImageNet.
In the first part of this work we present a performance analysis of different CNN layers for generating proposals. More specifically, similarly to BING \cite{BING}, we select a reduced set of window sizes and aspect ratios and slide them on each possible location of the feature map generated by a certain CNN layer. The relevance (or objectness) of the windows is learned using a linear classifier. As the proposal generation procedure should be fast, we base the feature aggregation for each candidate window on average pooling, which can be computed in constant time using integral images \cite{violaintegral}. 
From this analysis we see that there is not a single best layer for candidate windows generation. Instead we notice that deeper layers, having a more semantic representation, perform very well in recalling the objects with a reduced set of hypotheses. Unfortunately, as noticed also for other tasks \cite{hyper}, they provide a poor localization of the object due to their coarseness.
In contrast, earlier layers are better in accurately localizing the object of interest, but their recall is reduced as they do not represent strong object cues.
Thus, we conclude that, for a good window candidate generation, we should leverage multiple layers of the CNN. However, even with the very fast integral images for the feature extraction, evaluating all window locations at all feature layers is too expensive. Instead we propose a method based on a cascade starting from the last convolutional layer (layer 5) and going down with subsequent refinements until the initial layers of the net. As the flow of the cascade is inverse to the flow of the feature computation we call this approach an \textit{inverse cascade}. 
Also, as we start from a coarse spatial window resolution, and throughout the layers we select and spatially refine the window hypotheses until a reduced and spatially well localized set of hypotheses, we call our method \textit{coarse-to-fine inverse cascade}. An overview of our approach is illustrated in Fig.~\ref{fig:overview}. 
We evaluate the performance of the method in terms of recall vs. number of proposals as well as in terms of recall vs. object overlap. We show that in both evaluations the method is better than the current state of the art, and computationally very efficient.
However, the best of the method comes when it is associated with a CNN-based detector \cite{girshick15fastrcnn}. In this case the approach does not need to compute any feature, because it reuses the same features already computed by the CNN network for detection. Thus, we can execute the full detection pipeline efficiently.

In the next section, we describe related work.
Next, in section 3, we analyze the quality of different CNN layers for window proposal generation. 
Section 4 describes our inverse coarse-to-fine cascade. In section 5 we compare our method
with the state-of-the-art, both in terms of object proposal generation as in terms of object detection 
performance. Section 6 concludes the paper.

\section{Related work}
\label{sec:related}

\paragraph{Object proposal methods}

Object proposal generators aim at obtaining an accurate object localization with few object window hypotheses. These proposals can help object detection in two ways: searching objects in fewer locations to reduce the detector running time and/or using more sophisticated and expensive models to achieve better performance.

Object proposal methods can be grouped mainly in two approaches. 
The first measures objectness (\ie how likely an image window is an object) of densely sampled windows~\cite{objectness,BING,edgebox}.
Alexi \textit{et al}.~\cite{objectness} propose an objectness measure based on image saliency and other cues like color and edges. 
BING~\cite{BING} presents a very fast proposal generator by training a classifier on edge features, but it suffers from low localization accuracy.
Cracking BING \cite{crackBING} showed that the BING classifier has minimal impact on locating objects and without looking at the actual image a similar performance can be obtained.
Edgeboxes~\cite{edgebox} uses structural edges of \cite{structurededge}, a state-of-the-art contour detector, to
compute proposal scores in a sliding window fashion without any parameter learning. For a better localization it uses a final window refinement step. 
Like these methods, our approach densely samples hypotheses in a sliding window fashion. However, in contrast to them, we use a hierarchy of high-to-low level features extracted from a deep CNN which has proven to be effective for object detection~\cite{RCNN,regionlets}. 

An alternative approach to sliding-window methods is segmentation-based algorithms. This approach applies to the multiple levels of segmentation and then merge the generated segments in order to generate objects proposals~\cite{MCG,CPMC,randomprime,selectivesearch}. 
More specifically, selective search~\cite{selectivesearch} hierarchically aggregates multiple segmentations in a bottom-up greedy manner without involving any learning procedure, but based on low level cues, such as color and texture. 
Multiscale Combinatorial Grouping (MCG) ~\cite{MCG} extracts multiscale segmentations and merges them by using the edge strength in order to generate objects hypotheses.
Carreira \etal \cite{CPMC} propose to segment the object of interest based on graphcut. It produces segments from randomly generated seeds. As in selective search, each segment represents a proposal bounding box. 
Randomized Prim's \cite{randomprime} uses the same segmentation strategy as selective search. However, instead of merging the segments in a greedy manner it learns the probabilities for merging, and uses those to speed up the procedure. Geodesic object proposals \cite{Geodesic} are based on classifiers that place seeds for a geodesic distance transform on an over-segmented image.


\section{CNN layers for object proposals}
\label{sec:basic}
In this section we analyze the quality of the different layers of a CNN as features for window proposal generation. 

\subsection{Basic Approach}
\label{subsec:basic}

\paragraph{Sliding window}
Computing all possible boxes in a feature map of size $N \times N$ is in the order of $O(N^4)$ and therefore computationally unfeasible. Hence, similarly to \cite{BING} we select a set of window sizes that best cover the training data in terms of size and aspect ratio and use them in a sliding window fashion over the selected CNN layer. 
This approach is much faster than evaluating all possible windows and avoids to select windows with sizes or aspect ratios different from the training data and therefore probably false positives.

For the selection of the window sizes, we start with a pool of windows $W_{all}$ in different sizes and aspect ratios $W_{all}:\{\omega|\omega \in \mathbb{Z}^2, \mathbb{Z}=[1..20]\}$. 
It is important to select a set of window sizes that gives high recall (with \texttt{IoU}$>0.5$) and at the same time produces well localized proposals. 
To this end, for each window size, we compute its recall with different \texttt{IoU} thresholds and greedily pick one window size at a time that maximizes $\sum_{\alpha} recall(\texttt{IoU}>\alpha)$ over all the objects in the training set. Using this procedure, $50$ window sizes are selected for the sliding window procedure. 
In Fig.~\ref{fig:layer}(middle) we show the maximum recall that can be obtained with the selected window sizes, which is an upper bound of the achievable recall of our method.

\paragraph{Multiple scales}
Even though it is possible to cover all possible objects using a sliding window on a single scale of feature map, it is inefficient since by using a single scale the stride is fixed and defined by the feature map resolution. For an efficient sliding window the window stride should be proportional to the window size. 
Therefore, in all the experiments we evaluate our set of windows on multiple scales. For each scale, we resize the image such that $min(w,h)=s$ where $s\in \{227, 300, 400, 600\}$. Note that the first scale is the network original input size.

\paragraph{Pooling}
As the approach should be very fast we represent a window by average pooling of the convolutional features that are inside the window. As averaging is a linear operation, after computing the integral image, the features of any proposal window can be extracted in a constant time.
Let $f(x,y)$ be the specific channel of the feature map from a certain CNN layer and $F(x,y)$ its integral image.
Then, average pooling $avr$ of a box defined by the top left corner $a=(a_x,a_y)$ and the bottom right corner $b=(b_x,b_y)$ is obtained as:
\begin{equation}
	avr(a,b)=\frac{F(b_x,b_y)-F(a_x,b_y)-F(b_x,a_y)+F(a_x,a_y)}{(b_x-a_x)(b_y-a_y)}. 
	\label{equ:int}
\end{equation}
Thus, after computing the integral image, the average pooling of any box is obtained with a constant computational cost that corresponds to summing $4$ integral values and dividing by the area of the box.

\paragraph{Pyramid}
One of the main cues to detect general objects is the object boundaries.
Using an approach based on average pooling can dilute the importance of the object boundaries because it discards any geometrical information among features.
Therefore, to introduce more geometry to the description of a window we consider a spatial pyramid representation~\cite{Spatialpyramid}.
It consists of dividing the proposal window into a number of same size sub-windows (\eg $2\times2$), and for each one build a different representation.

\paragraph{Bias on size and aspect ratio}
Objects tend to appear at specific sizes and aspect ratios. Therefore we add in the feature representation 3 additional dimensions $(w,h,w\times h)$ where $w$ and $h$ are the width and height of window $\omega$ respectively. This can be considered as an explicit kernel which lets the SVM learn which object sizes can be covered in a specific scale.
For the final descriptor, we normalize the pooled features and size-related features separately with $l_2$ norm.

\paragraph{Classifier}
We train a linear classifier for each scale separately. For a specific scale, the classifier is trained with randomly selecting 10 regions per image that overlap the annotation bounding boxes more than 70\%, as positive training data and 50 regions per image that overlap less than 30\% with ground-truth objects as negative data. In all experiments we use a linear SVM \cite{LIBLINEAR} because of its simplicity and fast training. We did not test non-linear classifiers since they would be too slow for our approach.

\paragraph{Non-maximal suppression}
The ranked window proposals in each scale are finally reduced through a non-maximal suppression step. A window is removed if its \texttt{IoU} with a higher scored window is more than threshold $\alpha$. Varying the threshold $\alpha$ is a trade-off between recall and accurate localization. So, this threshold is directly related to the \texttt{IoU} criteria that is used for evaluation (see sec \ref{subsec:eval_basic}). By tuning $\alpha$, it is possible to maximize recall at arbitrary \texttt{IoU} of $\beta$. Particularly, in this work we define two variants of \methodname~namely \methodname50 and \methodname70 for maximizing recall at \texttt{IoU} of $\beta=0.5$ and $\beta=0.7$ respectively by fixing $\alpha$ to $\beta+0.05$ (like~\cite{edgebox}). In addition, to aggregate boxes from different scales, we use another non-maximal suppression, fixing $\alpha=\beta$.

\subsection{Evaluation}
\label{subsec:eval_basic}
\begin{figure*}
\begin{center}
\scalebox{0.8}
{
\begin{tabular}{ccc}
\includegraphics[width=0.32\linewidth]{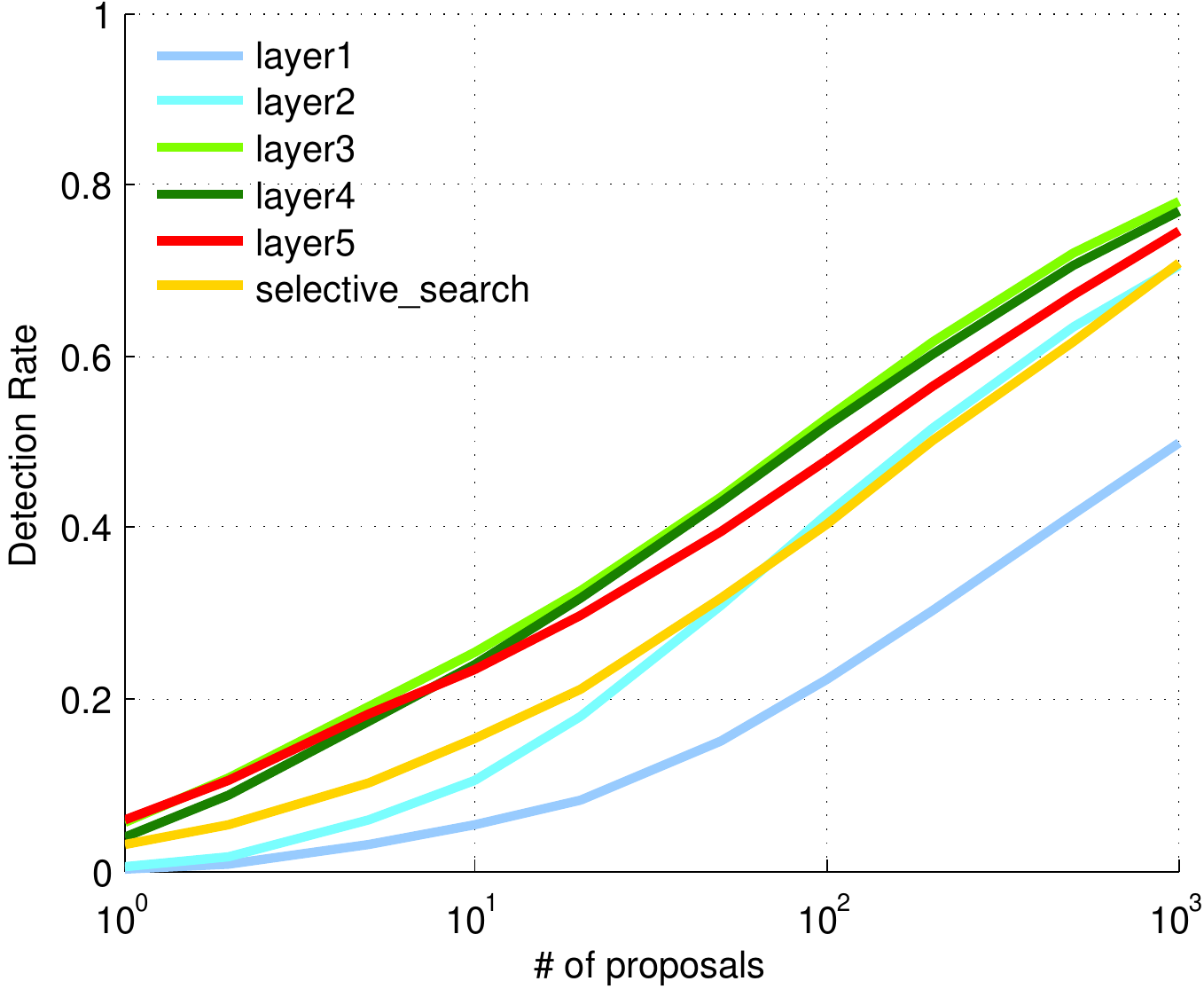}
&
\includegraphics[width=0.32\linewidth]{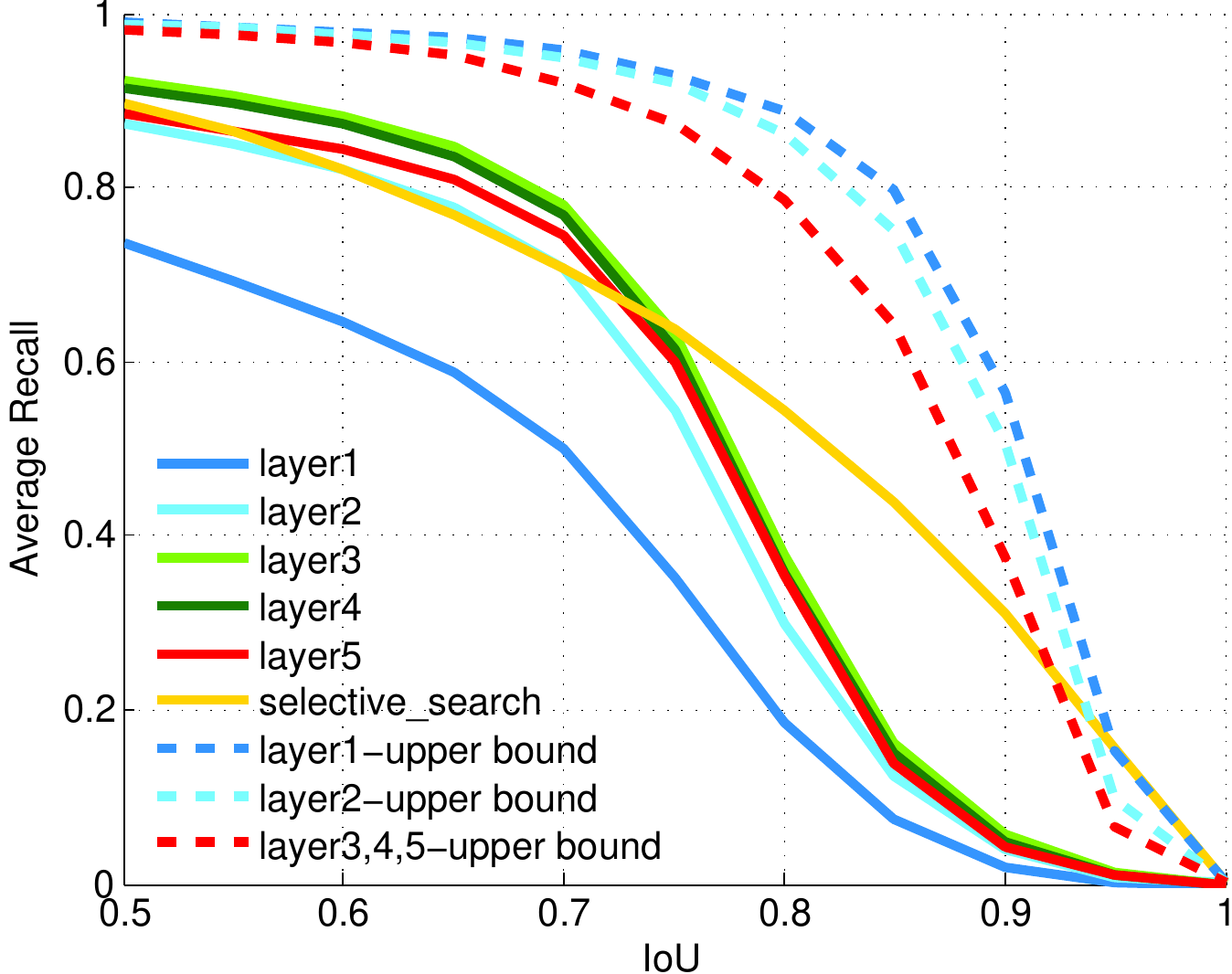}
&
\includegraphics[width=0.32\linewidth]{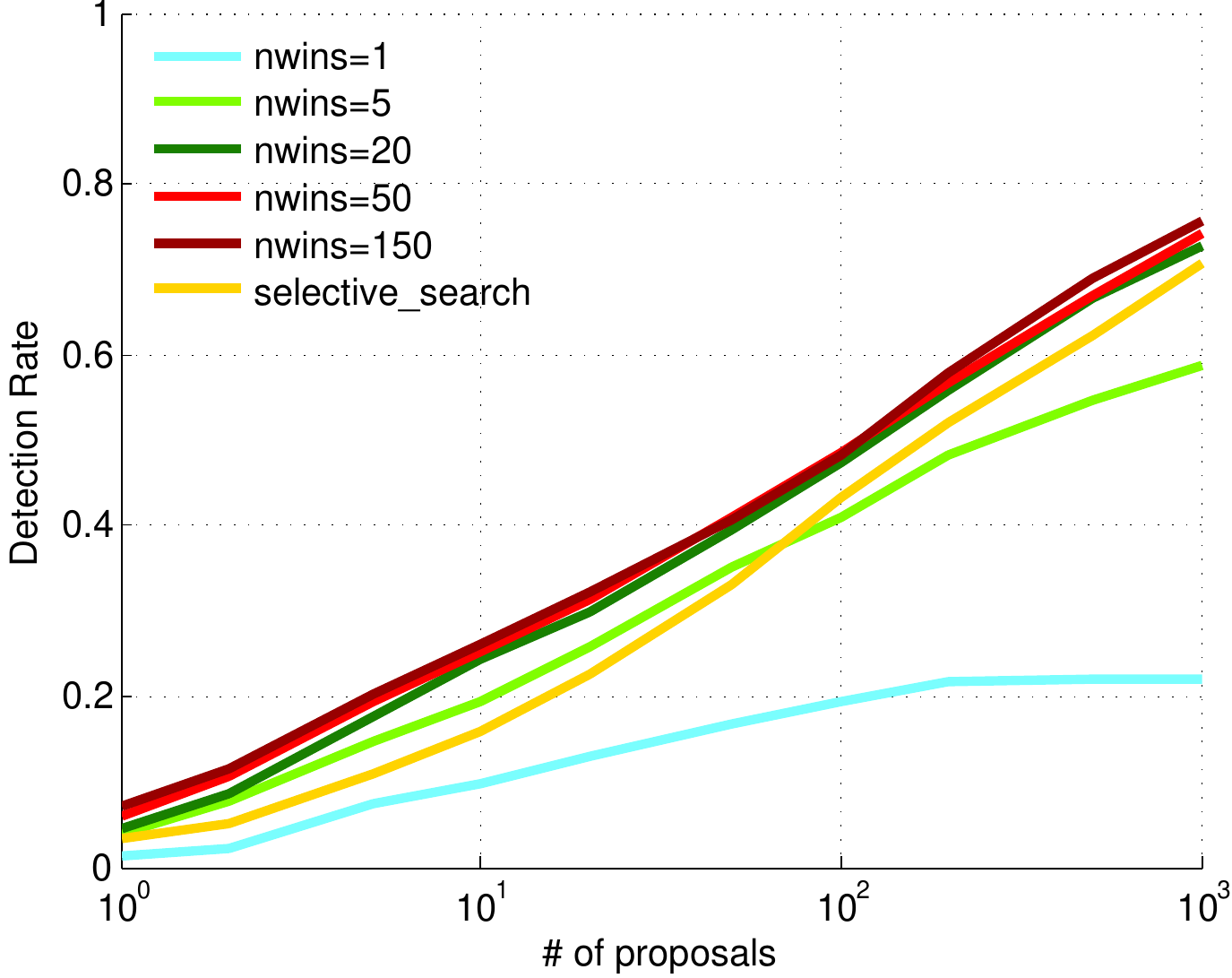}

\end{tabular}
}
\end{center}
\caption{({\bf{Left}}) Recall versus number of proposals for \texttt{IoU}=0.7. ({\bf{Middle}}) recall versus overlap for 1000 proposals for different layers. ({\bf{Right}}) Recall versus number of proposals at \texttt{IoU}=0.7 on layer 5 for different number of window sizes. All are reported on the PASCAL VOC 2007 test set.}
\label{fig:layer}
\end{figure*}

\begin{figure*}
\begin{center}
\scalebox{0.8}
{
\begin{tabular}{ccc}
\includegraphics[width=0.32\linewidth]{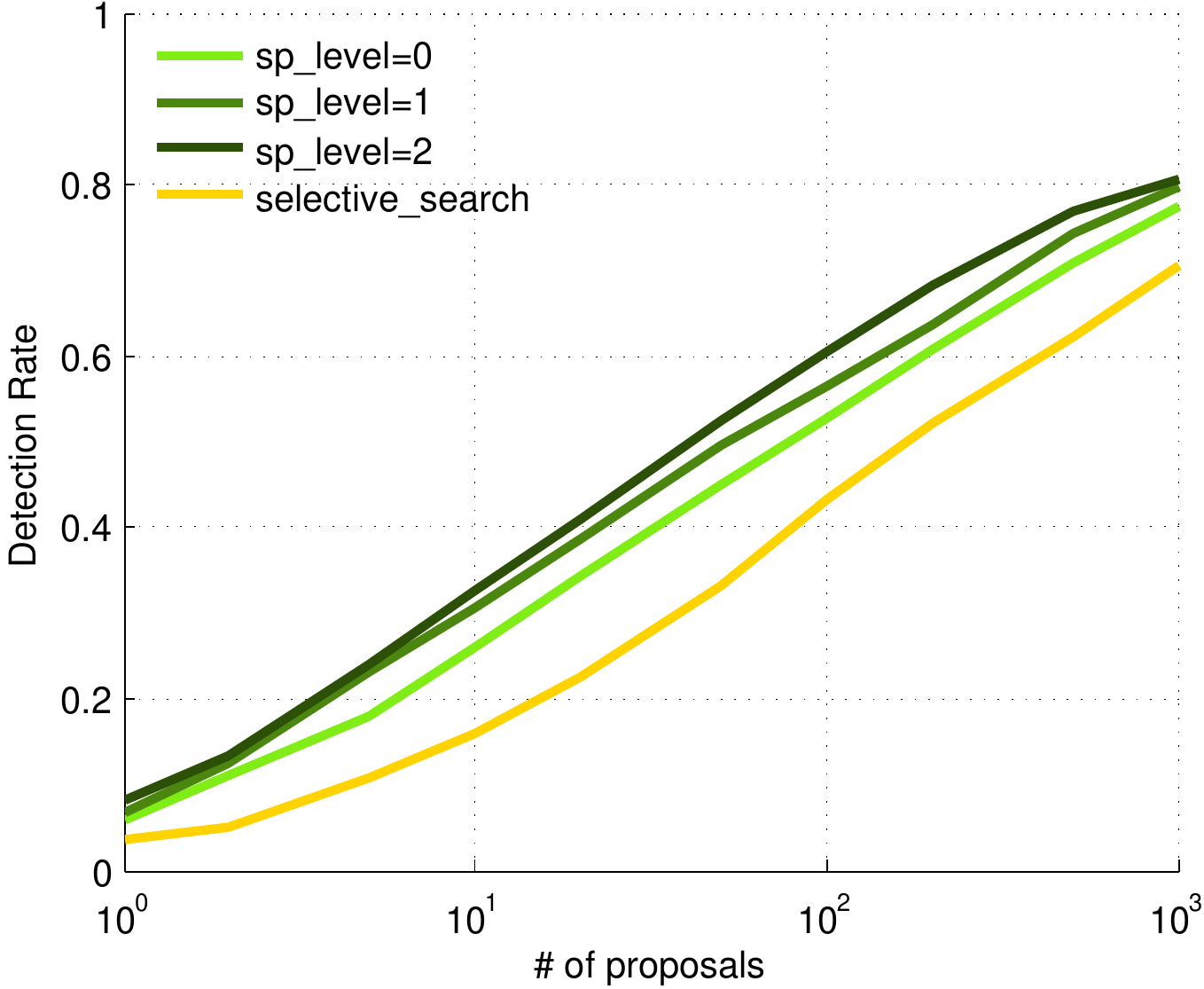}
&
\includegraphics[width=0.32\linewidth]{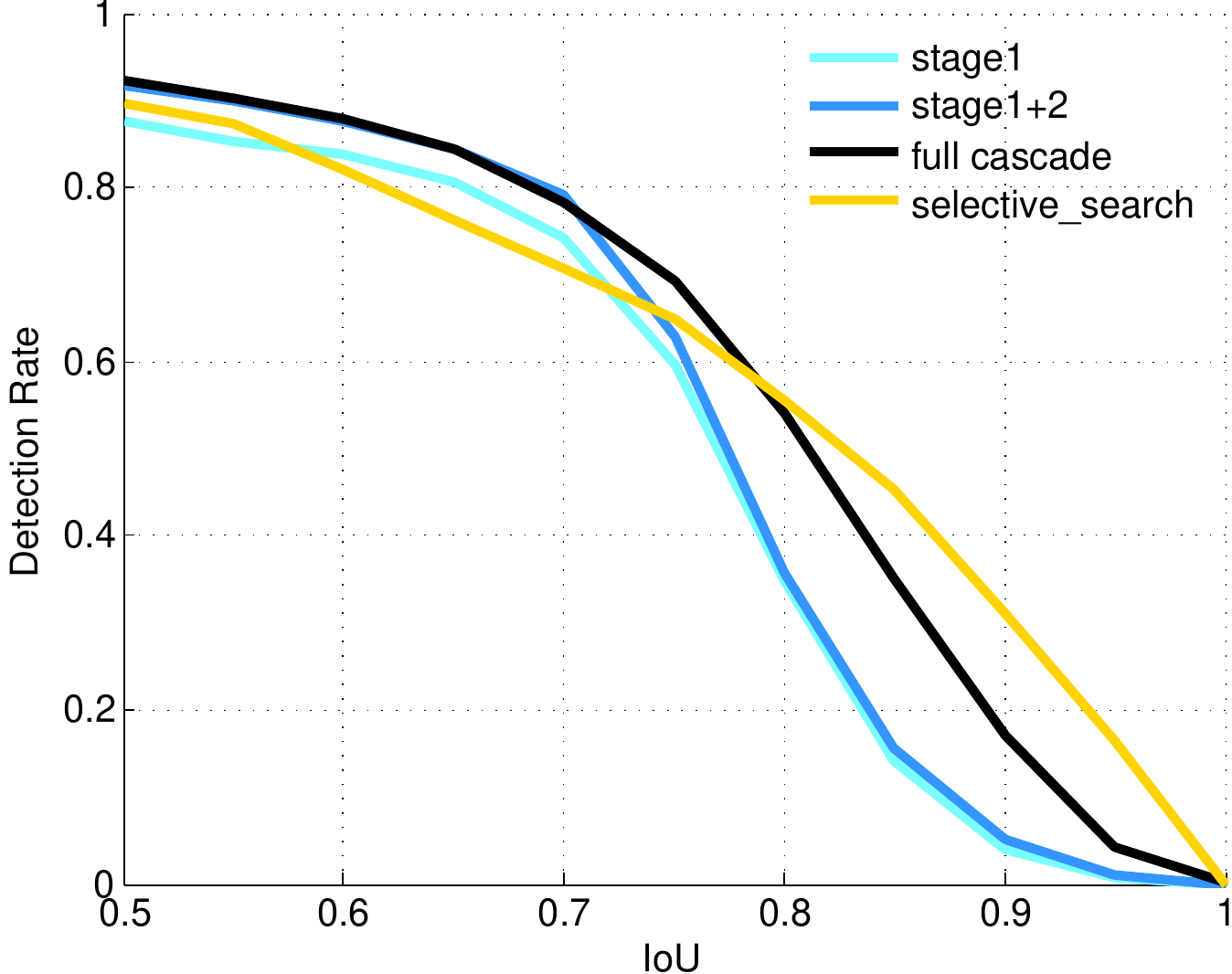}
&
\includegraphics[width=0.32\linewidth]{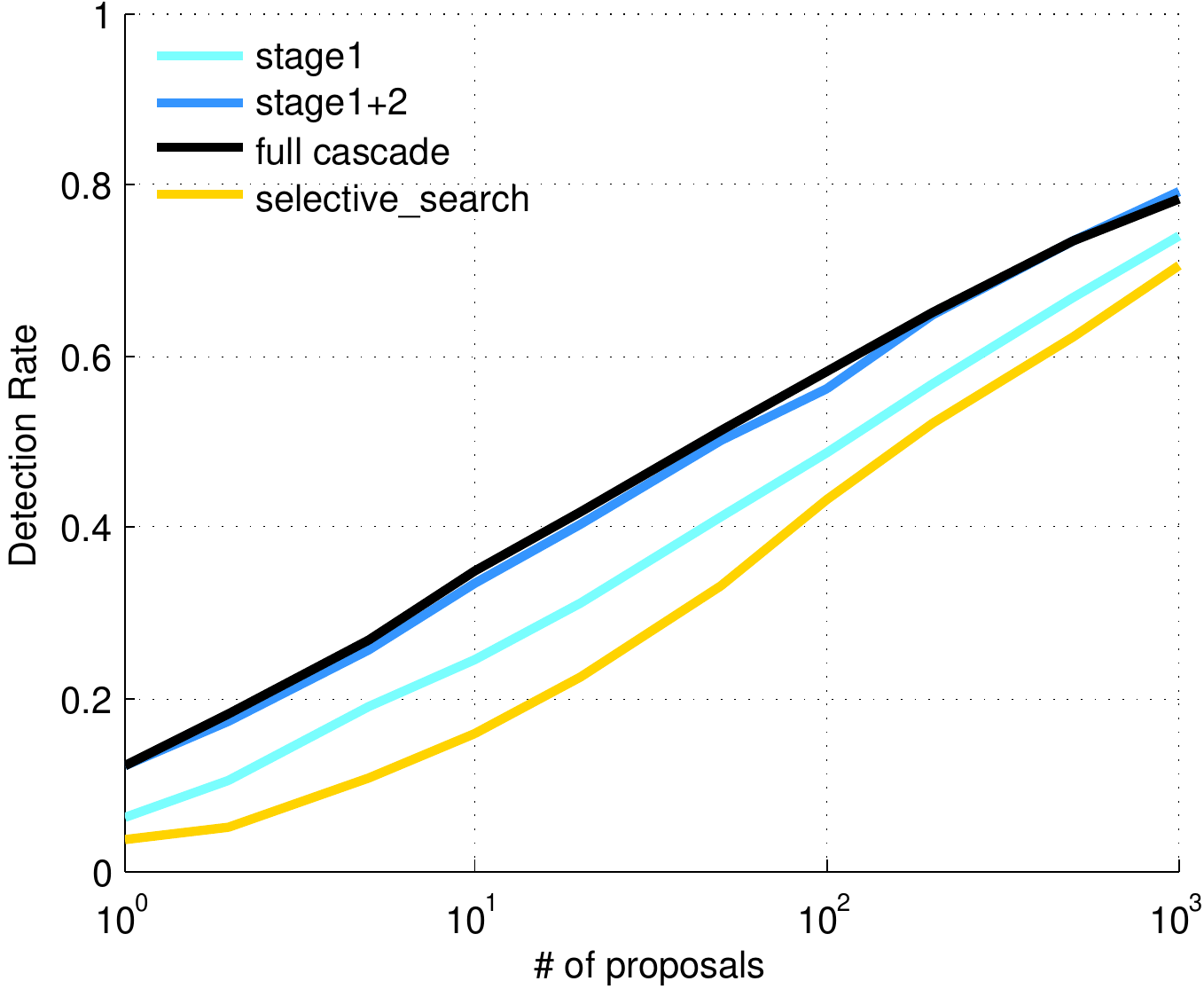}

\end{tabular}
}
\end{center}
\caption{({\bf{Left}}) Recall versus number of proposals in \texttt{IoU}=0.7 for different spatial pyramid levels ({\bf{Middle}}) Recall versus \texttt{IoU} for 1000 proposals for different stages of the cascade. ({\bf{Right}}) Recall versus number of proposals in \texttt{IoU}=0.7 for the different stages of the cascade. All are reported on the PASCAL VOC 2007 test set.}
\label{fig:var2}
\end{figure*}

\begin{table*}[t]
\centering
\scalebox{0.8}
{
\begin{tabular}{cccccccc}
	 Layer & Feature map size      & Recall(\#1000,0.5) & Max(0.5) & Recall(\#1000,0.8) & Max(0.8)\\
\hline
     5   &  $36\times52\times256$  & 88\% & 97\% & 36\%  & 70\%  \\
     4   &  $36\times52\times256$  & 91\% & 97\% & 36\%  & 79\%  \\
     3   &  $36\times52\times256$  & 92\% & 97\% & 38\%  & 79\%  \\
     2   &  $73\times105\times396$  & 87\% & 98\% & 29\%  & 86\%   \\
     1   &  $146\times210\times96$   & 73\% & 99\% & 18\%  & 89\%   \\
\end{tabular}
}
\caption{Characteristics and performance of the CNN layers. Feature map size is reported for an image of size $600 \times 860$. Recall(\#1000,$\beta$) is the recall of 1000 proposals for the overlap threshold $\beta$. Max($\beta$) is the maximum recall for the overlap threshold $\beta$ using our selected window sizes set. }
\label{table:cnn_char}
\end{table*}

For evaluating the quality of proposals, like previous works on object proposal generation, we focus on the PASCAL VOC 2007 dataset~\cite{pascal}.
PASCAL VOC 2007 includes 9,963 images with 20 object categories. 4,952 images are used for testing, while the remaining ones are used for training. 
We use two different evaluation metrics; the first is Detection Rate (or Recall) vs. Number of proposals. This measure indicates how many objects can be recalled for a certain number of proposals. 
We use Intersection over union (\texttt{IoU}) as evaluation criterion for measuring the quality of an object proposal $\omega$. \texttt{IoU} is defined as $|\frac{\omega\cap b}{\omega\cup b}|$ where $b$ is the ground truth object bounding box. Initially, an object was considered correctly recalled if at least one generated window had an \texttt{IoU} of $0.5$ with it, the same overlap used for evaluating the detection performance of a method. Unfortunately this measure is too lose because a detector, for working properly, needs also good alignment with the object \cite{Rodrigo14}. Thus we evaluate our method for an overlap of $0.7$ as well. 
We also evaluate recall vs. overlap for a fixed number of proposals. As shown in \cite{Rodrigo14}, the average recall obtained from this curve seems highly correlated with the performance of an object detector. 

In this section, we investigate the effect of different parameters of our method, namely the different convolutional layers, and the number of used windows. 

\paragraph{Layers}
We evaluate each convolutional layer (from 1 to 5) of Alexnet \cite{alexnet} using the sliding window settings explained above. We use Alexnet which is trained by Caffe toolbox \cite{caffe}. For sake of simplicity, we do not add spatial pyramids on top of pooled features in this set of experiments. As shown in Fig.~\ref{fig:layer}~({\bf left}) the top convolutional layers of the CNN perform better than the bottom ones. Also their computational cost is lower as their representation is coarser. Note this simple approach already performs on par or even better than the best proposal generator approaches. For instance, our approach at layer 3 for 100 proposals achieves a recall of $52\%$, whereas
selective search \cite{selectivesearch} obtains only $40\%$. This makes sense because the CNN features are specific for object classification and therefore can easily localize the object of interest.

However, this is only one side of the coin. If we compare the performance of the CNN layers for high overlap (see Fig.~\ref{fig:layer}~({\bf middle})), we see that segmentation based methods are much better \cite{selectivesearch, MCG}. For instance the recall of selective search for 1000 proposals at $0.8$ overlap is around $55\%$ whereas our at layer 3 is only $38\%$. This is due to the coarseness of the CNN feature maps that do not allow a precise bounding box alignment to the object. In contrast, lower levels of the net have a much finer resolution that can help to align better, but their encoding is not powerful enough to properly localize objects. In Fig.~\ref{fig:layer}~({\bf middle}) we also show the maximum recall for different overlap that a certain layer can attain with our selected sliding windows. In this case, the first layers of the net can recall many more objects with high overlap. This shows  that a problem of the higher layers of the CNN is the lack of a good spatial resolution. 

In this sense we could try to change the structure of the net in a way that the top layers still have high spatial resolution. However, this would be computationally expensive and, more importantly, it would not allow to reuse the same features used for detection. Instead, in the next section we propose an efficient way to leverage the expressiveness of the top layers of the net together with the better spatial resolution of the bottom layers.

\paragraph{Number of Sliding Windows}
In Fig.~\ref{fig:layer}~({\bf right}) we present the effect of a varying number of window sizes in the sliding window procedure for proposal generation. The windows are selected based on the greedy algorithm explained in Sec~\ref{subsec:basic}. 
As the number of used window sizes increases, we obtain a better recall at a price of a higher cost.
For the next experiments we will set the number of windows to $50$ because that is a good trade-off between speed and top performance.
The values in the figure refer to layer 5, however, similar behavior has been observed for the other layers.

\paragraph{Spatial Pyramid} 
We evaluate the effect of using a spatial pyramid pooling in Fig.~\ref{fig:var2}~({\bf left}). As expected, adding geometry improves the quality of the proposals. Moving from a pure average pooling representation (sp\_level=0) to a $2\times2$ pyramid (sp\_level=1) gives a gain that varies between $2$ and $4$ precent in terms of recall, depending on the number of proposals. Moving from the $2\times2$ pyramid to the $4\times4$ (sp\_level=2) gives a slightly lower gain. At $4\times4$ the gain does not saturate yet. However, as we aim at a fast approach, we also need to consider the computational cost, which is linear in the number of spatial bins used. Thus, the representation of a window with a $2\times2$ spatial pyramid is 5 times slower than a flat representation and the $4\times4$ pyramid is $21$ times slower. Thus, for our final representation we limit the use of the spatial pyramid to $2\times2$.



\section{Inverse Cascade}
\label{seec:hierarchy}

Even if the features used for our object proposals come without any additional computational cost (because they are needed for the detector), still a dense evaluation in a sliding window fashion over the different layers would be too expensive.
Instead here we leverage the structure of the CNN layers to obtain a method that combines in an efficient way the high recall of the top convolutional layers of a CNN, with the fine localization provided at the bottom layers of the net. In Table~\ref{table:cnn_char} we summarize the characteristics of each CNN layer.

We start the search with the top convolutional layers of the net, that have features well adapted to recognize objects, but are coarse, and then move to the bottom layers, that use simpler features but have a much finer spatial representation of the image (see Fig.~\ref{fig:overview}). As we go from a coarse to a fine representation of the image and we follow a flow that is exactly the opposite of how those features are computed we call this approach \textit{coarse-to-fine inverse cascade}.
We found that a cascade with 3 layers is an optimal trade-off between complexity of the method and gain obtained from the cascading strategy.

\paragraph{Stage 1: Dense Sliding Window on Layer 5}
The first stage of the cascade uses layer 5. As the feature representation is coarse, we can afford a dense sliding window approach with $50$ different window sizes collected as explained in Sec.~\ref{subsec:basic}. Even though a pyramid representation could further boost the performance, we do not use spatial binning at this stage to not increase the computational cost. We linearly map the window scores to $[0,1]$ such that the lowest and highest scores are mapped to $0$ and $1$ respectively. Afterwards we select the best $N_1=4000$ windows obtained from a non-maximum suppression algorithm with threshold $\beta+0.05$ in order to propagate them to the next stage.

\paragraph{Stage 2: Re-scoring Selected Windows on Layer 3}
In this stage, as we use a reduced set of windows, we can afford to spend more computation time per window. Therefore we add more geometry in the representation by encoding each window with a pyramid representation composed of two levels: $1\times1$ and $2\times2$. The proposal scores from this layer are again mapped to $[0,1]$. The final score for each proposal is obtained  multiplying the scores of both stages. Afterwards we apply a non-maximal suppression with overlap threshold $\beta+0.05$ and select the $3000$ best candidates. At the end of this stage, we aggregate the boxes from different scales using non-maximal suppression with threshold $\beta$ and select the $N_{desired}=1000$ best for refinement.

\paragraph{Stage 3: Local Refinement on Layer 2}
The main objective of this stage is to refine the localization obtained from the previous stage of the cascade. For this stage the best candidate is layer 2 because it has a higher resolution than upper layers and contains low-level information which is suitable for the refinement task. Specifically, we refine the $N_{desired}$ windows received from the previous stage using the procedure explained in~\cite{edgebox}. To this end, we train a structured random forest~\cite{structurededge} on the second layer of the convolutional features to estimate contours similarly to DeepContour~\cite{deepcontour}. After computing the edgemap, a greedy iterative search tries to maximize the score of a
proposal over different locations and aspect ratios using the scoring function used in~\cite{edgebox}. 
It is worth mentioning that since our contour detector is based on the CNN-features, we again do not need to extract any extra features for this step.

\subsection{Evaluation}
\label{subsec:eval_hierarchy}

\begin{table*}[t]
\centering
\begin{tabular}{cccccccc}
	Stage & Layer & input candidates & Method       & Pyramid & NMS    & Total time per image\\
\hline
     1    &   5   &  $\sim$80.000  	& Slid. Window 	& $1$       & Yes &  0.30s\\
     2    &   3   &  4.000  		& Re-scoring    & $1+2\times2$       & Yes &  0.25s\\
     3    &   2   &  1.000  		& Refinement    & -       & No  &  0.20s \\
\end{tabular}
\caption{Characteristics of the stages of our inverse cascade (NMS: non maximum suppression).}
\label{table:stages}
\end{table*}
We discuss the performance of the inverse cascade stage by stage in terms of both computational cost and performance. A summary of the computational cost of each stage is given in Table~\ref{table:stages}.
The entire cascade has a computational cost of $0.75$ seconds, which is the composition of $0.3$ , $0.25$ and $0.2$ for the first, second and third stage respectively. Note the first stage is very fast because even if we use a dense sliding window approach, with the integral image and without any pyramid level the cost of evaluating each window is very low. 

As shown in Fig.~\ref{fig:var2}~({\bf middle and right}), 
the second stage is complementary to the first and employed with a $2\times2$ pyramid improves the recall of the cascade by $5\%$. However, this boost is valid only up to an overlap of $0.75$. After this point the contribution of the second stage is negligible. This is due to the coarse resolution of layer $5$ and $3$ that do not allow a precise overlap of the candidate windows with the ground truth object bounding boxes. We found that, for our task, layer 3 and 4 have a very similar performance (Recall@1000 is 79\% in both cases) and adding the latter in the pipeline could not help in improving performance (Recall@1000 is still 79\%).

As shown in \cite{Rodrigo14}, for a good detection performance, not only the recall is important, but also a good alignment of the candidates as well. At stage $3$ we improve the alignment without performing any further selection of windows; instead we refine the proposals generated by the previous stages by aligning them to the edges of the object. In our experiments for contour detection we observed that layer 1 of CNN did not provide as good performance as layer 2 (0.61 vs. 0.72 AP on BSDS dataset~\cite{arbelaez2011contour}) so we choose second layer of network for this task. Fig.~\ref{fig:var2}~({\bf middle}) shows this indeed improves the recall for high  \texttt{IoU} values (above 0.7).


\section{Experiments}
\label{sec:comparison}

In this section we compare the quality of the proposed \methodname~with state-of-the-art object proposals. 
In section~\ref{subsec:results} we compare the quality of our \methodname~in terms of recall and localization accuracy for PASCAL VOC 2007.

Then, in section~\ref{subsec:detectionRes} detection results are reported for PASCAL VOC 2007 \cite{pascal} using Fast-RCNN~\cite{girshick15fastrcnn}. Finally in section ~\ref{subsec:generalization}, we evaluate the generalization performance of \methodname~on unseen categories.


\subsection{Comparison with state-of-the-art}
\label{subsec:results}
\begin{figure*}[t]
\begin{minipage}{.67\textwidth}
\begin{center}
\begin{tabular}{c@{}c@{}c@{}c}
\includegraphics[width=0.46\linewidth]{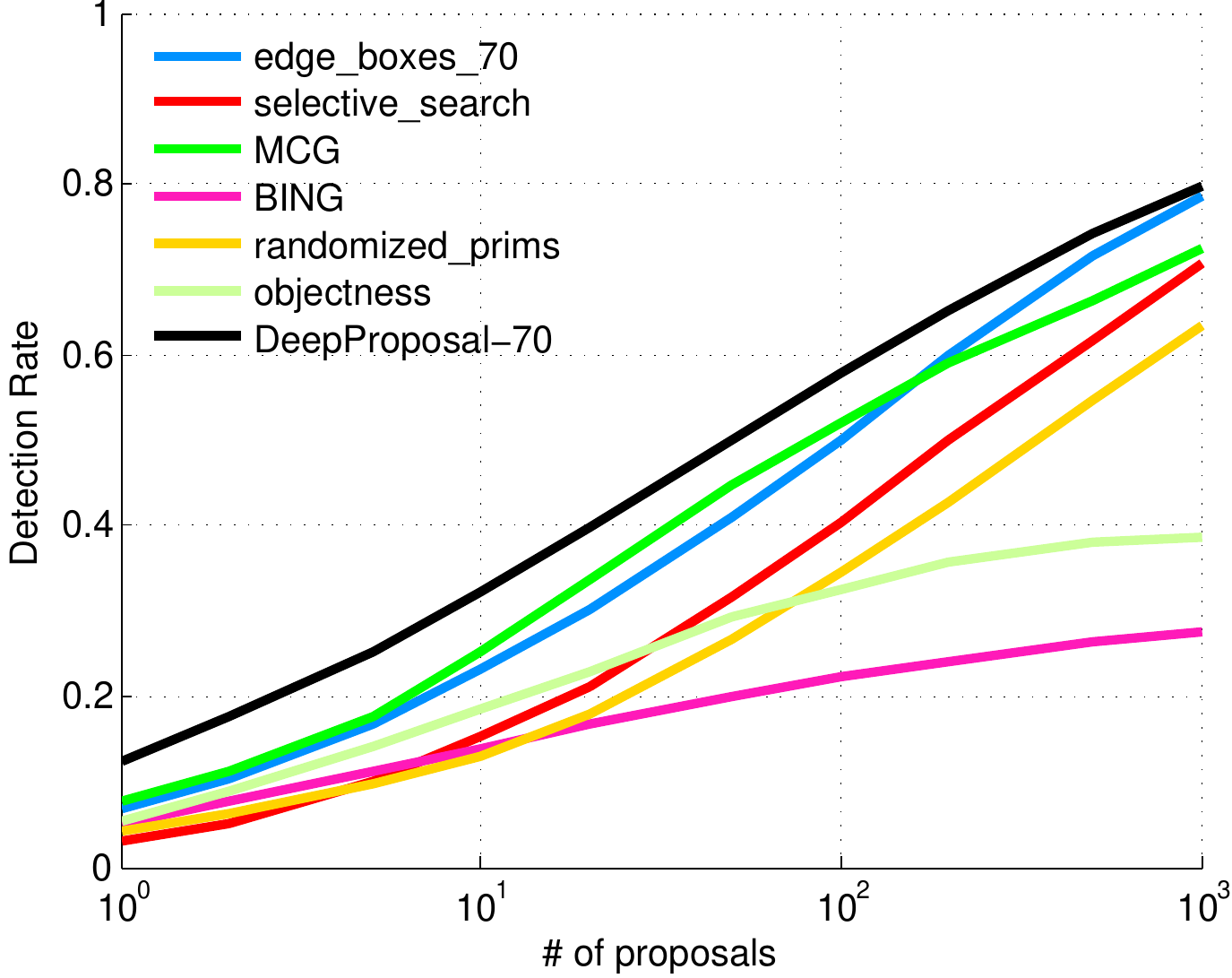}
&
\includegraphics[width=0.46\linewidth]{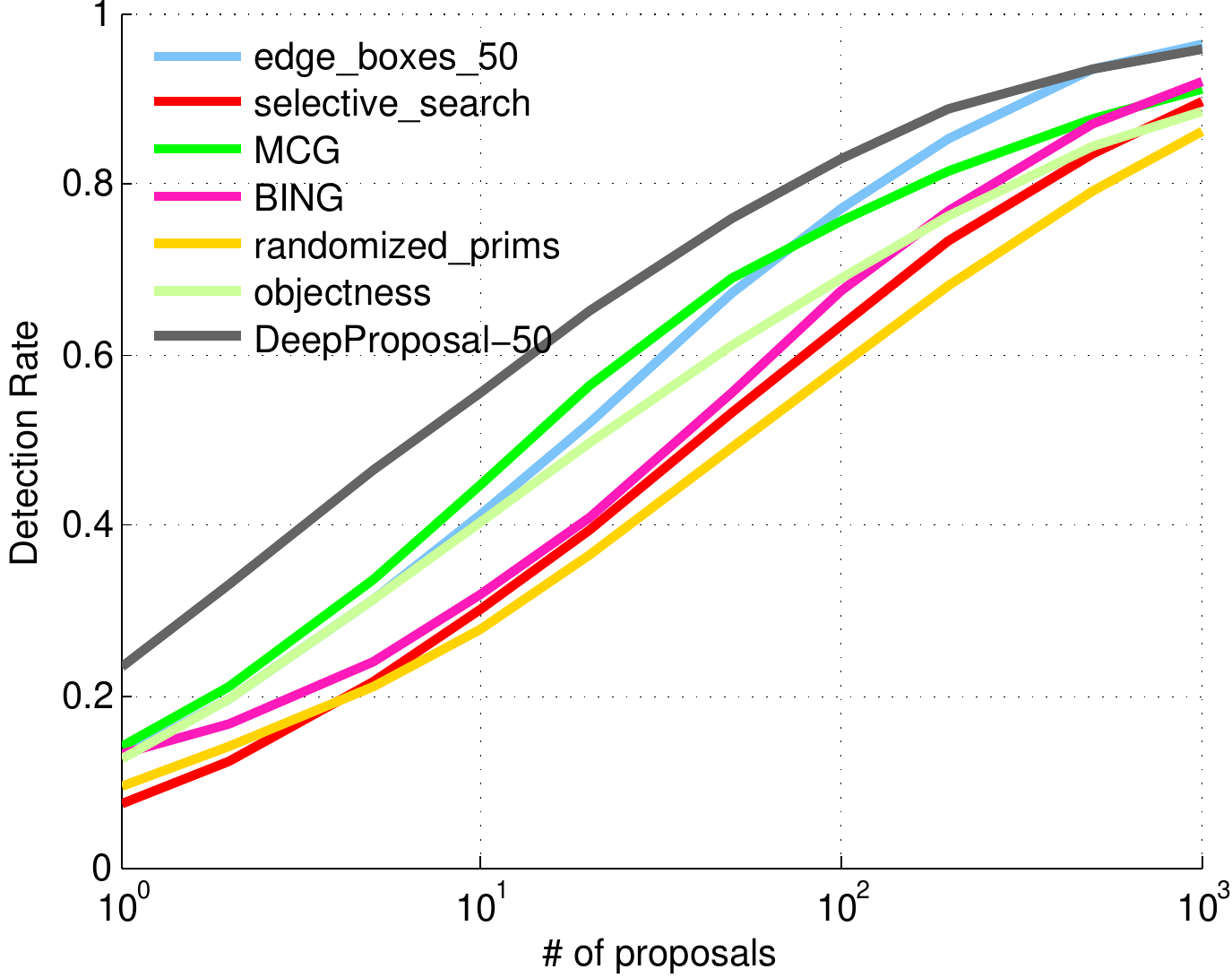}
\end{tabular}
\captionof{figure}{Recall versus number of proposals on the PASCAL VOC 2007 test set for ({\bf{left}}) IoU threshold 0.5 and ({\bf{right}})IoU threshold 0.7.}
\label{fig:nprop}
\end{center}
\end{minipage}%
\begin{minipage}{.31\textwidth}
\begin{center}
{
{\includegraphics[width=0.95\linewidth]{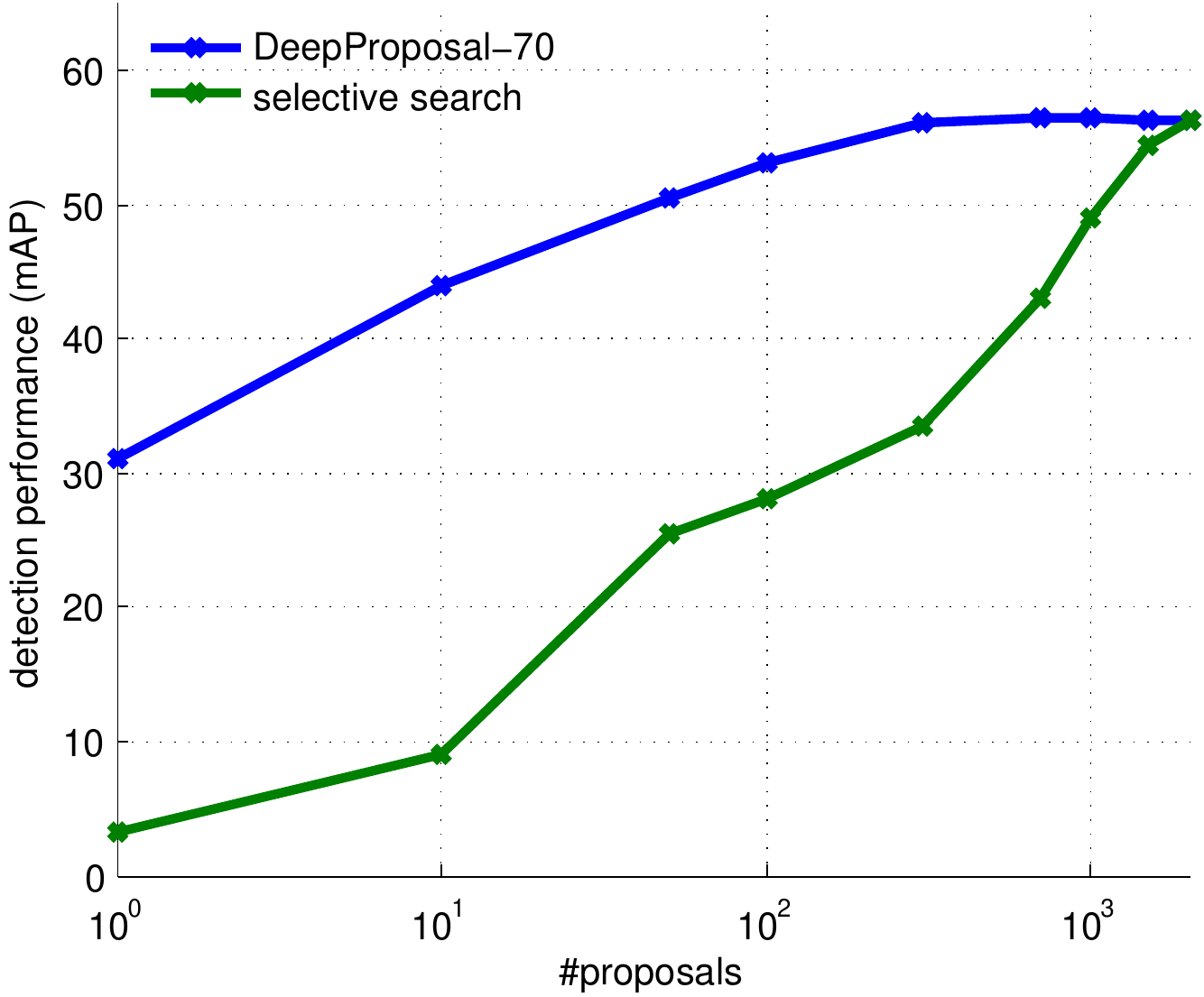}} 
}
\end{center}
\vspace{-0.2cm}
\captionof{figure}{Detection results on PASCAL VOC 2007.}
\label{fig:det_res}
\end{minipage}%
\end{figure*}

\begin{figure*}[t]
\begin{center}
\scalebox{0.9}
{
\begin{tabular}{c@{}c@{}c@{}c}
\includegraphics[width=0.32\linewidth]{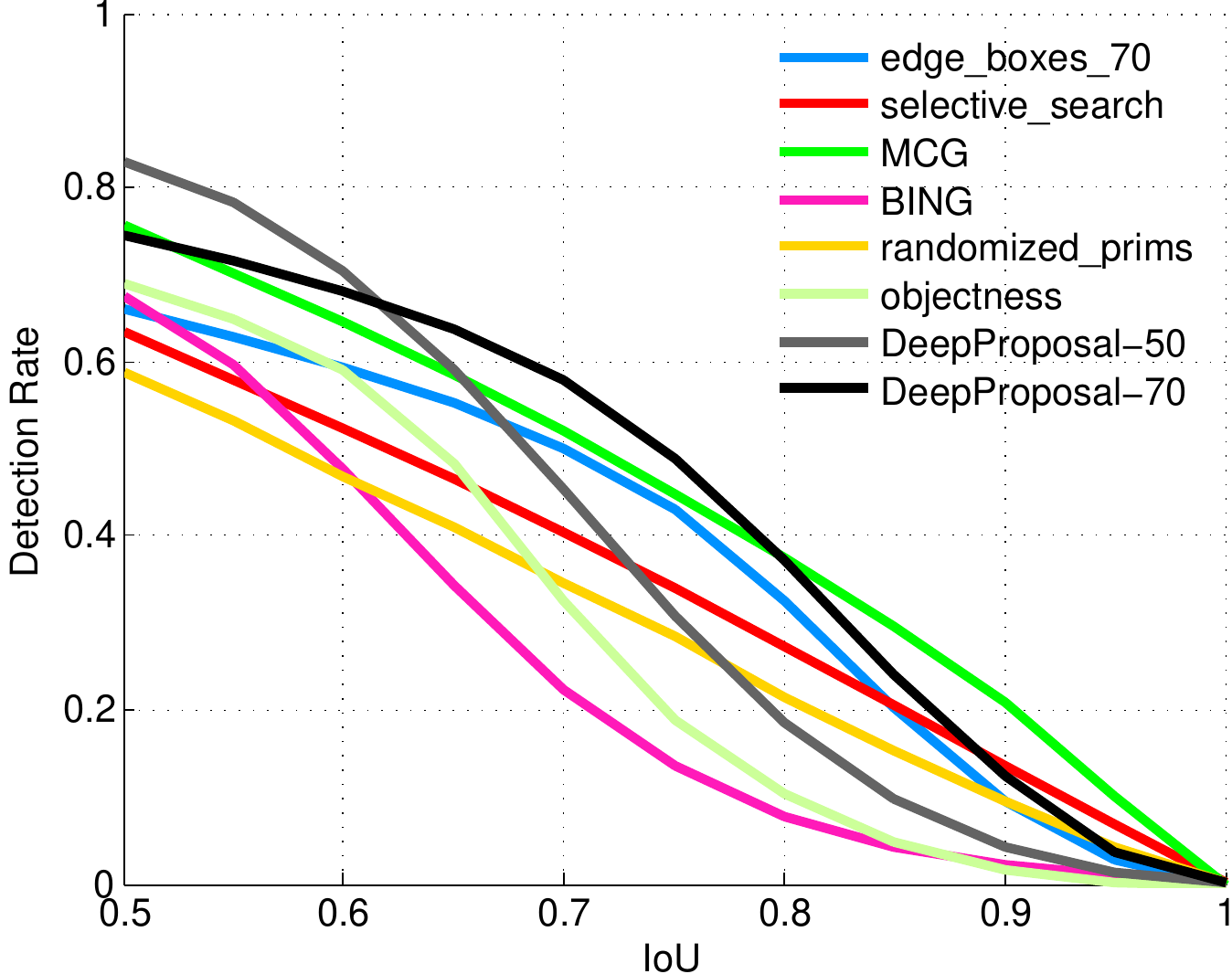}
&
\includegraphics[width=0.32\linewidth]{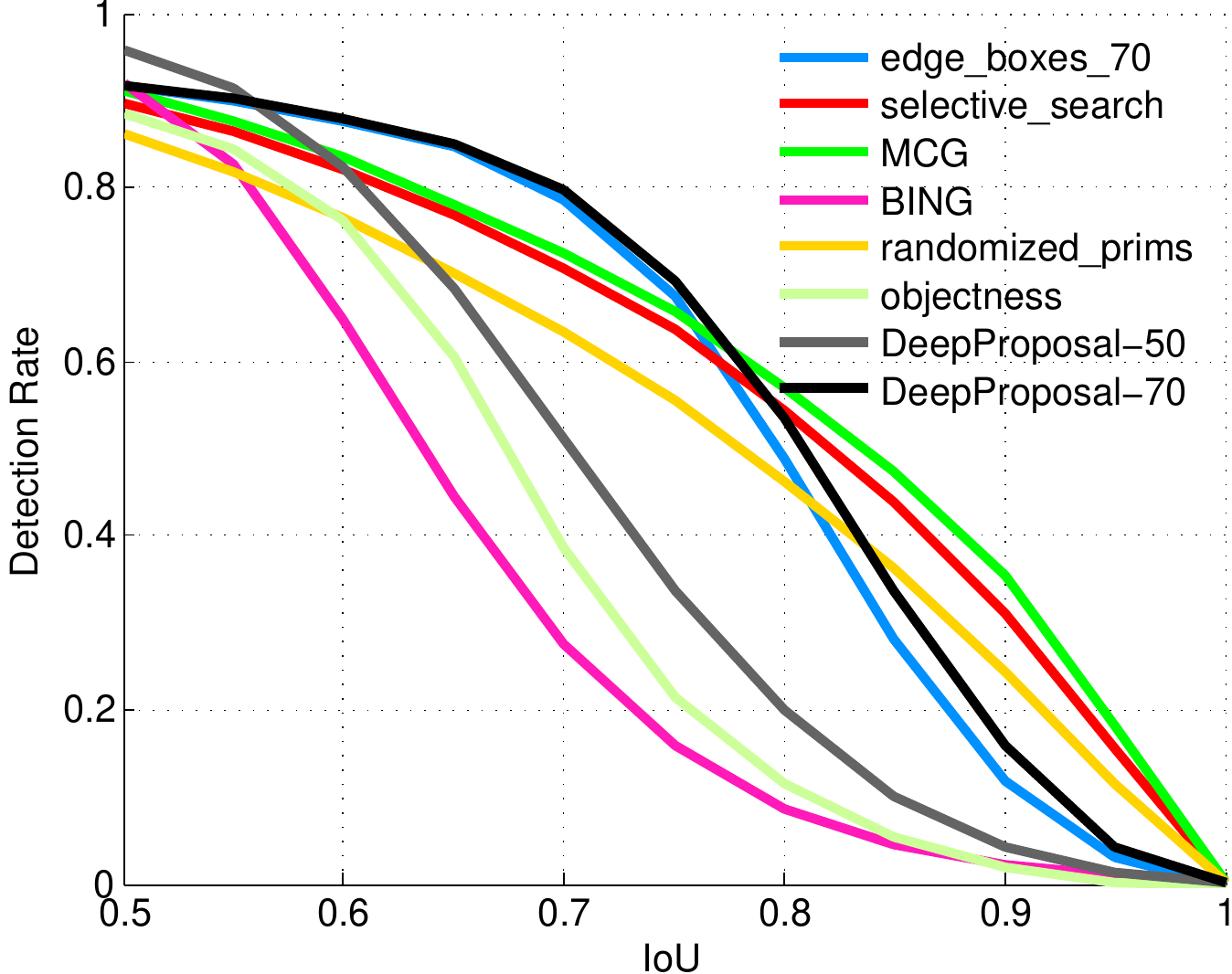}
&
\includegraphics[width=0.32\linewidth]{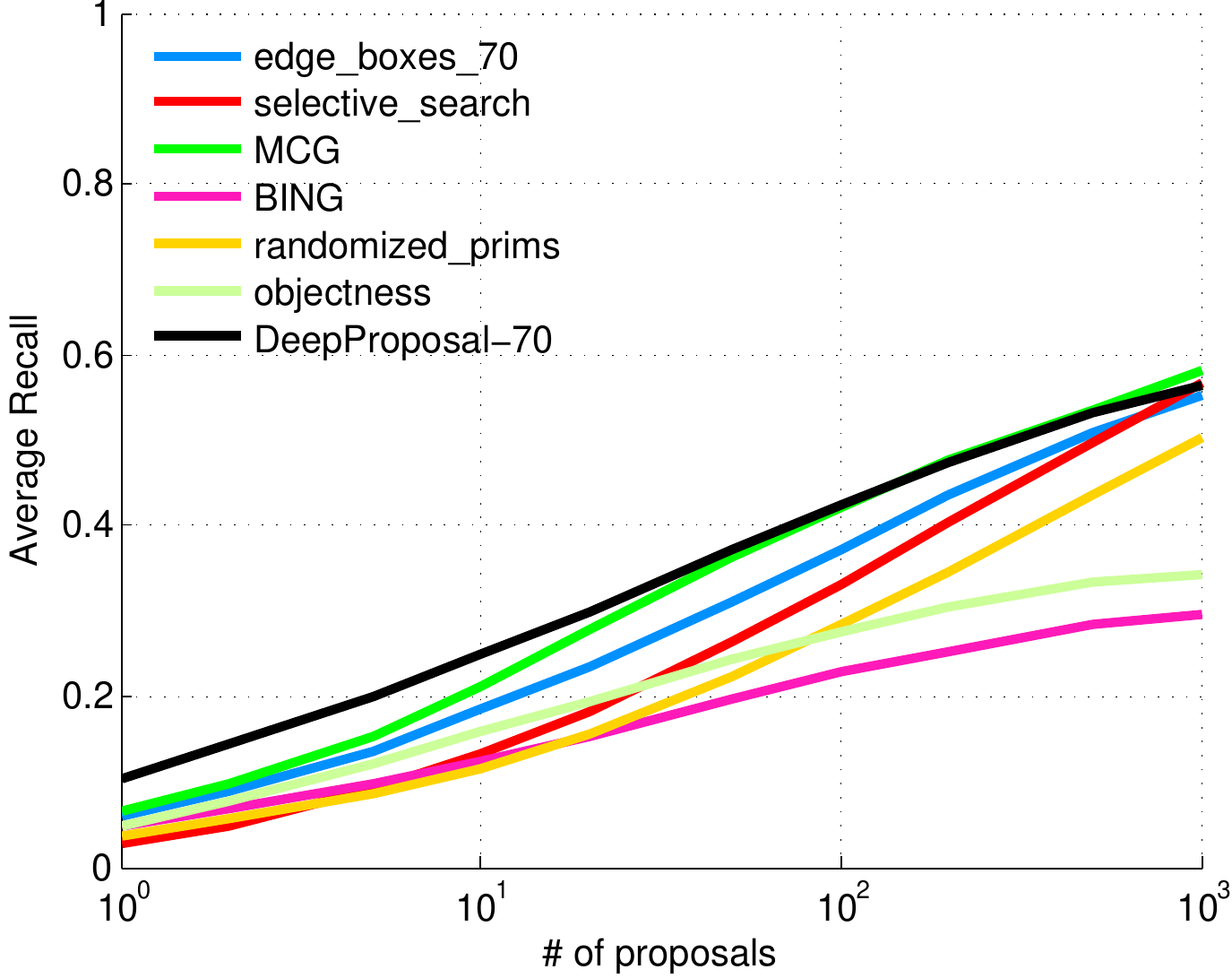}
\end{tabular}
}
\caption{Recall versus IoU threshold on the PASCAL VOC 2007 test set for ({\bf{left}}) 100 proposal windows and ({\bf{middle}})1000 proposal windows. ({\bf{right}}) Average Recall between [0.5,1] \texttt{IoU} on the PASCAL VOC 2007 test set}
\label{fig:iou}
\end{center}
\end{figure*}

In this section, we compare our \methodname~against well-known, state-of-the-art object proposal generators. Fig.~\ref{fig:nprop} and Fig.~\ref{fig:iou} show the recall with changing number of the object proposals or \texttt{IoU} threshold respectively. These curves reveal how \methodname~performs on varying \texttt{IoU}. From Fig.~\ref{fig:nprop}, we can conclude that, even with a small number of windows, \methodname~can achieve higher recall for any \texttt{IoU} threshold.
Methods like BING \cite{BING} and objectness \cite{objectness} are  providing high recall only at 
\texttt{IoU} = 0.5 because they are tuned for \texttt{IoU} of 0.5. 

When comparing results over a variety of \texttt{IoU} thresholds (Fig.~\ref{fig:iou}), we can see \methodname~achieves competitive
or higher recall and produces large enough number of proposal boxes. In table~\ref{table:cmpr} we evaluate the quality of proposals generated by all methods in a different way. 
Achieving 75\% recall with \texttt{IoU} value 0.7 would be possible with $540$ windows of \methodname, $800$ of Edge boxes, $1400$ using selective search proposals and $3000$ of Randomized Prim's windows \cite{randomprime}. 
Other methods are not comparable with these values of recall and \texttt{IoU} threshold.

Figure~\ref{fig:iou} shows the curves related to recall over changing amount of \texttt{IoU} with 100 and 1000 proposals. Again, \methodname~obtains good results in this test as well. The hand crafted segmentation based methods like selective search 
and MCG have good recall rate at higher \texttt{IoU} values. Instead \methodname~perform better in the range of \texttt{IoU} = [0.6, 0.8] which is desirable in practice and playing an important
role in object detectors performance \cite{Rodrigo14}.

Figure~\ref{fig:iou} ({\bf{right}}) shows average recall(AR) versus number of proposals for different methods. For a specific number of proposals, AR measures the proposal quality across \texttt{IoU} of [0.5, 1]. Hosang \etal.~\cite{Rodrigo14} shows that AR correlates well with detection performance. Using this criteria, \methodname~are on par or better than other methods with 700 or fewer boxes but with more boxes, selective search and Edgeboxes performs better.

The runtime tests for our proposed method and the others are available in Table~\ref{table:cmpr}. Since our approach is using the CNN features which are used 
by state-of-the-art object detectors like RCNN \cite{RCNN} and SppNet \cite{sppnet} and does not need any extra cues and features, we can consider just running time of
our algorithm without CNN extraction time\footnote{If features have to be (re)computed, that adds 0.7 sec. extra computation time on a low-end GPU.}. \methodname~takes 0.75 second on CPU and 0.4 second on a regular GPU which is just a bit slower than Edgeboxes. The fastest method is BING which has the lowest accuracy in any evaluation. The other methods which are segmentation based, take considerably more time.

\begin{table*}[t]
\centering
\scalebox{0.9}
{
\begin{tabular}{l|*{6}{c}}
			& AUC & N@25\% & N@50\% & N@75\% & Recall  & Time \\
\hline
\hline
BING\cite{BING} 			& .19 & 292 & -   & -    & 29\% & \textbf{.2s} \\
Objectness\cite{objectness}		& .26 & 27  & - & -  & 39\% & 3s\\
Rand. Prim's\cite{randomprime}            & .30 & 42  & 349 & 3023 & 71\% & 1s  \\
Selective Search\cite{selectivesearch}        & .34 & 28  & 199 & 1434 & 79\% & 10s \\
Edge boxes 70\cite{edgebox}		& .42 & 12  & 108 & 800  & \textbf{84\%} & .3s\\
MCG\cite{MCG}    			& .42 & 9   & 81  & 1363 & 78\% & 30s \\
\methodname70 		& \textbf{.49} & \textbf{5}   & \textbf{50}  & \textbf{540}  & 82\% & .75s \\
\end{tabular}
}
\caption{Our method compared to other methods for \texttt{IoU} threshold of 0.7. AUC is the area under recall vs. \texttt{IoU} curve for 1000 proposals. N@25\%, N@50\%, N@75\% are the number of proposals needed to achieve a recall of 25\%, 50\% and 75\% respectively. For reporting Recall, at most 2000 boxes are used. The runtimes for other method were obtained from~\cite{Rodrigo14}.}
\label{table:cmpr}
\end{table*}

\subsection{Object detection Performance}
\label{subsec:detectionRes}

In the previous experiments we evaluate our proposal generator with different metrics and show that it is among the best methods in all of them. However, we believe the best way to evaluate the usefulness of the generated proposals is a direct evaluation of the detector performance particularly that recently it has become clear (see \cite{Rodrigo14}) that an object proposal method with high recall at $0.5$ IoU does not automatically lead to a good detector.

The most performing detectors at the momet are: RCNN~\cite{RCNN}, SppNet~\cite{sppnet} and fast-RCNN~\cite{girshick15fastrcnn}. All are based on CNN features and use object proposals for detecting the object of interest.
The first uses the window proposals to crop the corresponding regions of the image, compute the CNN features and obtain a classification score for each region. This approach is slow and takes around $10$ sec on a high-end GPU and more than $50$ sec on the GPU used for our experiments.

SppNet and fast-RCNN instead compute the CNN features only once, on the entire image. Then, the proposals are used to select the sub-regions of the feature maps from where to pull the features.
This allows this approach to be much faster. With these approaches then, we can also reuse the CNN features needed for the generation of the proposal so that the complete detection pipeline can be executed without any pre-computed component roughly in $1$ second on our GPU. 

We compare the detection performance of our \methodname70 with selective search. Both methods are evaluated training a detector using the corresponding proposals, so that detector and proposal generator are matched and the comparison is fair. The training is conducted using fast-RCNN on PASCAL VOC 2007. In Fig.~\ref{fig:det_res} we report the detector mean average precision on the PASCAL VOC 2007 test data for different number of used proposals. As expected the difference between the two approaches is quite relevant and it appears mostly in a regime with low number of proposals. For instance, when using 100 proposals selective search obtains a mean average precision of $28.1$, while our proposals already reach $53.2$. Also, our proposals reach almost the top performance with only $300$ bounding boxes, while selective search needs more than $2000$ boxes to reach its best performance. This is an important factor when seeking for maximum speed. We believe that this different behavior is due to the fact that our method is supervised to select good object candidates, whereas selective search is not.

Using SppNet fine-tuned for selective search, we obtain a mAP of 52.2 with \methodname~which is lower than 54.5 of the selective search. Similar behavior has been reported for other methods since the model is trained on selective search~\cite{Rodrigo14}.

Another advantage of our approach, being based on learning, is that it can focus on specific classes. In this sense we train a special version of \methodname~for cars, where the positive training samples are collected only from car instances.
In this setting the performance of the car detector improves from $57.6\%$ to $60.4\%$ using SppNet. Thus, in this scenario, our proposals can also be use to improve a detector performance.

\subsection{Generalization to unseen categories}
\label{subsec:generalization}
We evaluate the generalization capability of our approach on Microsoft COCO dataset \cite{coco}. The evaluation of the approach has been done by learning either from the 20 classes from VOC07 or from 5, 10, 20, 40, 80 randomly sampled from COCO. 
When the \methodname~is trained by only 5 classes, the recall at $0.5$ IoU with 1000 proposals is slightly reduced ($56\%$).
With more classes, either using VOC07 or COCO, recall remains stable around $59\%$ - $60\%$. 
This shows that the method can generalize over all classes. We believe this is due to the simplicity of the classifier (average pooling on CNN features) that avoids overfitting specific classes. 
Note that in this case our recall is slightly lower than the Selective Search with 1000 proposals ($63\%$).
This is probably due to the presence of very small objects that our system is not tuned for.
These results on COCO demonstrate that our proposed method
is capable to generalize learnt objectness beyond the training categories.


\section{Conclusion}
\label{sec:conclusion}
\methodname, the method that is proposed in this paper is a way to produce object proposal windows, based on convolutional neural network activation features
as used in state-of-the-art object detectors. We provide an algorithm to use one kind of feature for both localization and detection, which makes the
object detectors needless of any extra features or different method to extract possible locations of objects. By employing an efficient coarse to fine cascade on multiple layers of CNN features, we have a framework of objectness measurement that acts strongly on objects locations and our method can find reasonable accurate proposals, fast. Source code will be made available online.

\subsection*{Acknowledgements}
This work was supported by DBOF PhD scholarship, KU Leuven CAMETRON project and FWO project ``Monitoring of Abnormal Activity with Camera Systems''.

{\small
\bibliographystyle{ieee}
\bibliography{refes}
}

\end{document}